\documentclass[11pt]{article}

\usepackage[final]{acl}

\usepackage{times}
\usepackage{latexsym}

\usepackage[T1]{fontenc}

\usepackage[utf8]{inputenc}

\usepackage{microtype}

\usepackage{inconsolata}

\usepackage{graphicx}

\usepackage{subfigure}
\usepackage{booktabs} 

\usepackage{array}
\usepackage{amsmath}
\usepackage{amssymb}
\usepackage{mathtools}
\usepackage{amsthm}

\usepackage{bm}
\usepackage{enumitem}
\usepackage{multirow}
\usepackage[table]{xcolor}

\usepackage[capitalize,noabbrev]{cleveref}

\theoremstyle{plain}

\theoremstyle{definition}

\theoremstyle{remark}

\usepackage[textsize=tiny]{todonotes}
\newlength\savewidth
\newcommand{\shline}{\noalign{\global\savewidth\arrayrulewidth
  \global\arrayrulewidth 1pt}
  \hline
  \noalign{\global\arrayrulewidth\savewidth}}

\title{EntropyPrune: Matrix Entropy Guided Visual Token Pruning for Multimodal Large Language Models}

\author{
    {\bf Yahong Wang}$^{1}$ \quad
    {\bf Juncheng Wu}$^{2}$ \quad
    {\bf Zhangkai Ni}$^{1,}$\footnotemark[2] \quad
    {\bf Chengmei Yang}$^{1}$ \quad
    {\bf Yihang Liu}$^{1}$ \\
    {\bf Longzhen Yang}$^{1}$ \quad
    {\bf Yuyin Zhou}$^{2}$ \quad
    {\bf Ying Wen}$^{3}$ \quad
    {\bf Lianghua He}$^{1,4,}$\footnotemark[2] \\
    \textsuperscript{1}Tongji University \quad
    \textsuperscript{2}University of California, Santa Cruz  \\
    \textsuperscript{3}East China Normal University \quad 
    \textsuperscript{4}Shanghai Eye Disease Prevention and Treatment Center \\
}

\begin{document}
\maketitle
{
\renewcommand{\thefootnote}{\fnsymbol{footnote}}
\footnotetext[2]{Corresponding authors.}
}

\begin{abstract}
Multimodal large language models (MLLMs) incur substantial inference cost due to the processing of hundreds of visual tokens per image. 
Although token pruning has proven effective for accelerating inference, determining when and where to prune remains largely heuristic.
Existing approaches typically rely on static, empirically selected layers, which limit interpretability and transferability across models. 
In this work, we introduce a matrix-entropy perspective and identify an \textbf{``Entropy Collapse Layer''} (ECL), where the information content of visual representations exhibits a sharp and consistent drop, which provides a principled criterion for selecting the pruning stage. 
Building on this observation, we propose \textbf{EntropyPrune}, a novel matrix-entropy-guided token pruning framework that quantifies the information value of individual visual tokens and prunes redundant ones without relying on attention maps. 
Moreover, to enable efficient computation, we exploit the spectral equivalence of dual Gram matrices, reducing the complexity of entropy computation and yielding up to a 64$\times$ theoretical speedup. 
Extensive experiments on diverse multimodal benchmarks demonstrate that EntropyPrune consistently outperforms established competitive pruning methods in both accuracy and efficiency. 
On LLaVA-1.5-7B, our method achieves a 57.7\% reduction in FLOPs while preserving 98.8\% of the original performance.
Furthermore, EntropyPrune generalizes effectively to high-resolution and video-based models, highlighting the strong robustness and scalability in practical MLLM acceleration~\footnote{Our code is available at \url{https://github.com/YahongWang1/EntropyPrune}}.
\end{abstract}

\section{Introduction}

Multimodal large language models (MLLMs)~\citep{bai2025qwen2,chen2025sft,liu2023visual,chen2024internvl} have recently achieved remarkable progress on a wide range of visual understanding and reasoning tasks~\citep{lu2022learn,fu2024mmecomprehensiveevaluationbenchmark,singh2019towards}. 
By integrating a visual encoder with a large language model, these systems enable flexible and general-purpose multimodal reasoning~\citep{vicuna2023,2023internlm,bai2023qwentechnicalreport}. 
However, existing MLLMs tend to represent images using a large number of visual tokens, leading to excessive input sequence lengths and high computational overhead.
For example, LLaVA-1.5~\citep{liu2024improved} represents each image using 576 visual tokens, while Qwen2.5-VL~\citep{bai2025qwen2} adopts a resolution-adaptive strategy that frequently produces several thousand tokens for high-resolution inputs. 

\definecolor{textred}{RGB}{255,0,0}
\definecolor{textgreen}{RGB}{0, 204, 0}

\begin{figure}[t]
\begin{center}
\centerline{\includegraphics[width=1.0\linewidth]{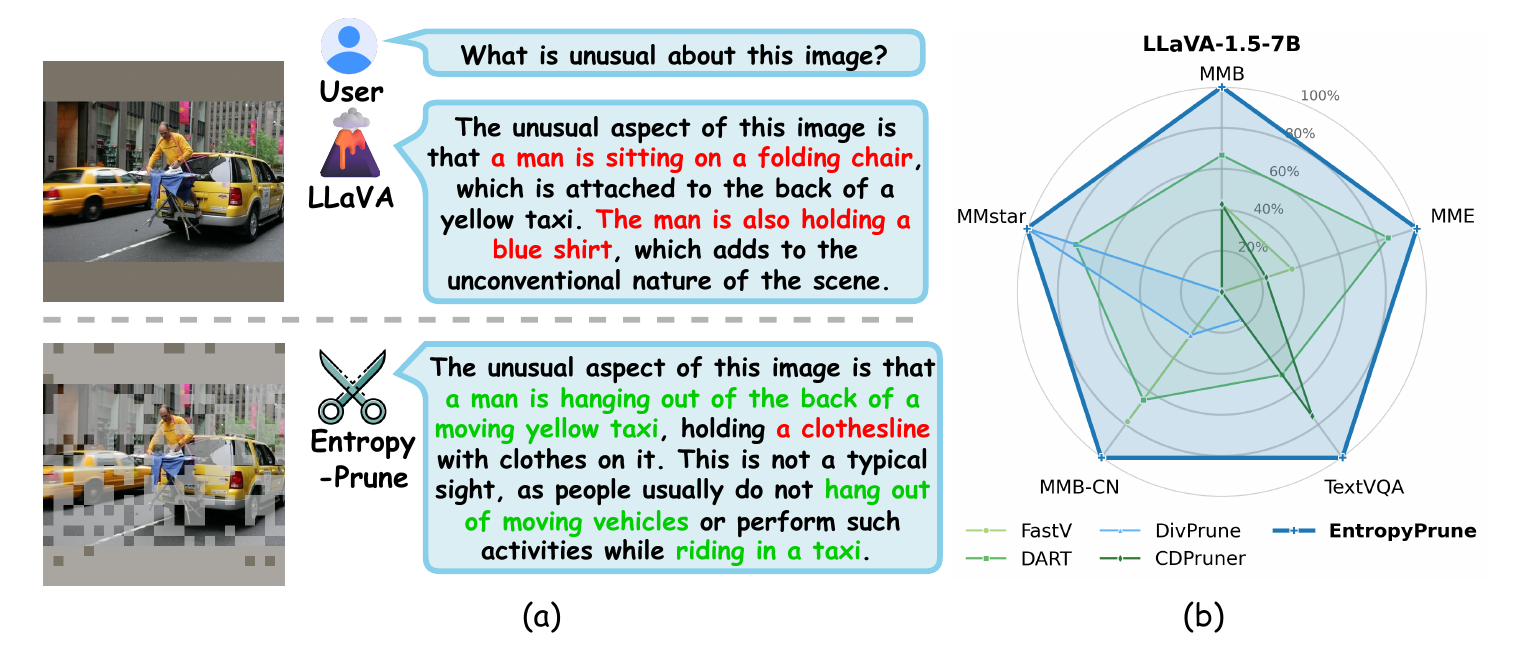}}
\vspace{-3mm}
\caption{
(a) \textbf{Comparison between vanilla LLaVA-1.5-7B and EntropyPrune.} 
Correct answers are highlighted in \textbf{\textcolor{textgreen}{green}}, while hallucinations are marked in \textbf{\textcolor{textred}{red}}. 
By removing low-information tokens, EntropyPrune encourages the model to concentrate on more critical details
(\textit{e.g.}, the person's state and the car's color).
(b) \textbf{Performance comparison.} 
The radial-axis visualization of min-max normalized scores shows that EntropyPrune consistently outperforms competitive methods, including FastV, DART, DivPrune, and CDPruner.
}
\vspace{-33pt}
\label{fig:teaser}
\end{center}
\end{figure}

\begin{figure*}[t]
\begin{center}
\centerline{\includegraphics[width=1.0\linewidth]{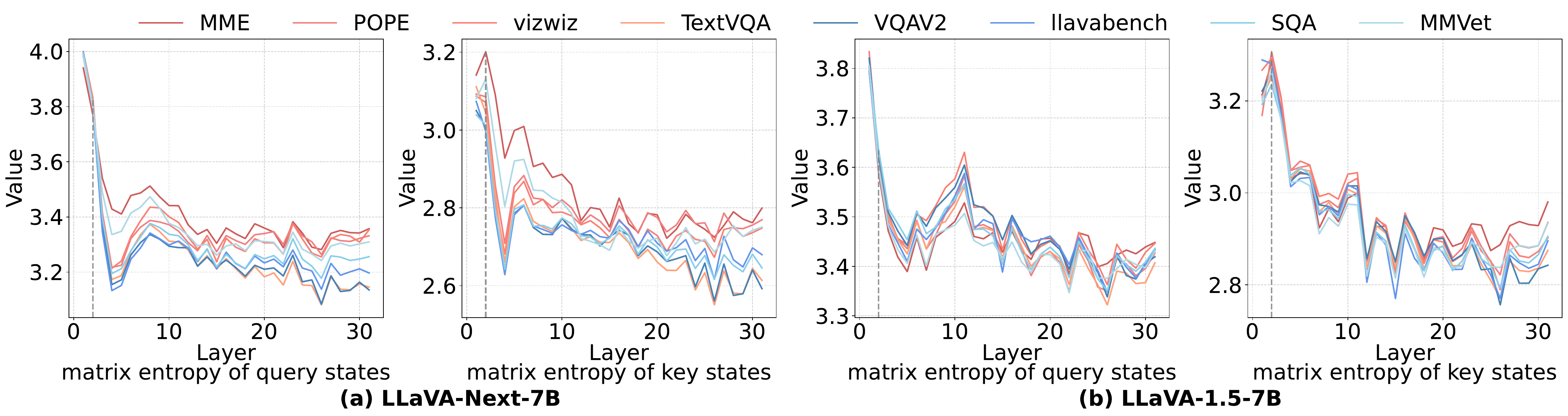}}
\vspace{-4mm}
\caption{
Layer-wise matrix entropy of visual tokens (query and key states) in LLaVA-1.5-7B and LLaVA-Next-7B across eight datasets.
A consistent layer-wise trend is observed across different datasets, with a precipitous entropy drop after the second layer.
}
\vspace{-30pt}
\label{fig:layer_matrix_entropy_llava}
\end{center}
\end{figure*}

To improve the efficiency of MLLM deployment, it is crucial to reduce the number of visual tokens.
Existing training-free visual token pruning strategies can be broadly categorized into two groups. 
Attention-based methods estimate token importance from attention weights and discard tokens with low scores~\citep{chen2024image,zhang2024sparsevlm,ye2025fit,liu2024multi,zhao2025stitch}. 
Diversity-based methods, in contrast, remove redundant tokens by measuring feature similarity~\citep{bolya2022tome,alvar2025divprune,wen2025stop,wang2025folder,zhang2026beyond}. 
These methods have shown promising results in reducing inference cost while preserving performance.
However, a fundamental question remains largely unexplored: \textbf{at which layers should pruning be applied?}
Most existing methods rely on statically selected pruning layers obtained via empirical tuning or grid search. 
Such heuristic choices lack interpretability, are model-dependent, and fail to reflect the intrinsic information flow of multimodal representations.

In this work, we revisit token pruning by analyzing the layer-wise information density of MLLMs from an information-theoretic perspective.
Inspired by matrix entropy theory~\citep{giraldo2014measures}, we characterize the information content of visual token representations using the entropy of trace-normalized covariance matrices. 
This formulation connects naturally to the von Neumann entropy in quantum information theory~\citep{vonNeumann1955mathematical}, enabling a principled quantification of the informational capacity of visual tokens. 
We analyze the layer-wise entropy dynamics of visual tokens in LLaVA-1.5-7B and LLaVA-NeXT-7B~\citep{liu2024llavanext} using randomly sampled inputs from eight datasets, including SQA~\citep{lu2022learn}, MME~\citep{fu2024mmecomprehensiveevaluationbenchmark}, TextVQA~\citep{singh2019towards}, POPE~\citep{li2023evaluating}, vizwiz~\citep{gurari2018vizwiz}, llavabench~\citep{liu2023visual}, MMVet~\citep{yu2024mm}, and VQAV2~\citep{balanced_vqa_v2}. 

As illustrated in \cref{fig:layer_matrix_entropy_llava}, our analysis reveals a consistent layer-wise pattern of matrix entropy across diverse datasets, independent of query or key states.
Here, we define the \textbf{``Entropy Collapse Layer''} (ECL) as the layer at which a sharp reduction in matrix entropy is observed. 
Specifically, the matrix entropy of the query and key states exhibits a precipitous decline after a specific layer (\textit{e.g.}, after the second layer in LLaVA-1.5-7B and LLaVA-Next-7B).
This sharp drop indicates a sudden reduction in the information carried by visual tokens, identifying this layer as an interpretable indicator for initiating pruning, unlike prior methods that rely on manual layer selection.

Based on this insight, we propose \textbf{EntropyPrune}, a novel training-free token pruning method that leverages entropy collapse to adaptively guide the pruning process. 
We apply matrix entropy to the token level to precisely quantify the information content of individual visual tokens.
Tokens with high entropy are retained, while low-information ones are removed, without relying on attention maps.
Nevertheless, naively computing matrix entropy requires eigendecomposition with cubic complexity. 
To alleviate this bottleneck, we introduce a spectral acceleration strategy that exploits the duality of Gram matrices. 
This optimization yields a theoretical $64\times$ speedup, making our method computationally feasible in practice.
Extensive experiments demonstrate that EntropyPrune substantially reduces computational overhead with negligible impact on model performance.
As shown in Figure~\ref{fig:teaser}, our method produces more accurate responses with fewer hallucinations than the vanilla model. 
On LLaVA-1.5-7B, EntropyPrune removes \textbf{66.7\%} of visual tokens and reduces inference FLOPs by \textbf{57.7\%}, while retaining \textbf{98.8\%} of the original performance without any additional training. 
In addition, EntropyPrune generalizes effectively to long-sequence scenarios, including high-resolution inputs and video understanding, highlighting its robustness and scalability.

In summary, our contributions are as follows:
\begin{itemize}
\item 
We identify a consistent \textbf{entropy collapse} phenomenon in MLLMs and introduce the \textbf{Entropy Collapse Layer} (ECL) as an interpretable criterion for pruning layer selection.
\item 
We propose \textbf{EntropyPrune}, a training-free token pruning framework that ranks visual tokens using matrix entropy and incorporates an efficient spectral acceleration strategy based on dual Gram matrices, achieving a theoretical $64\times$ speedup.
\item 
We conduct extensive evaluations on diverse image and video benchmarks, demonstrating competitive quality--efficiency trade-offs compared with advanced pruning methods.
\end{itemize}

\section{Related Work}
\label{sec:relatedWork}

\subsection{Visual Token Pruning in MLLMs}
To alleviate the computational burden introduced by the lengthy visual sequences in MLLMs, visual token pruning has emerged as a promising acceleration strategy. Existing training-free pruning approaches generally fall into two paradigms: \emph{attention-based} and \emph{diversity-based}. \textbf{Attention-based methods} treat attention weights as a proxy for token importance, thereby selecting the most important visual tokens~\citep{chen2024image,zhang2024sparsevlm,zhao2025stitch,ju2024turbo,xing2024pyramiddrop}. 
However, a critical limitation of attention-based methods is their dependence on explicit attention maps, making them incompatible with FlashAttention~\cite{dao2022flashattention, dao2023flashattention2}. 
Conversely, \textbf{Diversity-based methods} aim to eliminate redundancy by calculating similarity between visual tokens, thereby offering better compatibility with efficient optimizations~\citep{bolya2022tome,alvar2025divprune,wen2025stop,zhang2026beyond,li2025todrevisualtokenpruning}. 
Although these paradigms strike a reasonable trade-off between inference speed and accuracy, they often rely on empirical chosen layers to initiate pruning, which lack interpretability and adaptability.

\subsection{Matrix Entropy}

Matrix Entropy quantifies the intrinsic information content of data representations directly through the spectral properties of kernel matrices, bypassing explicit probability density estimation~\citep{giraldo2014measures}.
UNComp~\citep{xiong-etal-2025-uncomp} leverages matrix entropy to quantify the redundancy within the Key-Value (KV) cache of Large Language Models.
Most recently, HalfV~\citep{shi2026halfv} utilizes matrix entropy to analyze cross-modal token interactions across layers and designs a two-stage pruning framework.
Advancing this paradigm, our work identifies a precipitous drop in visual token information at a specific layer, defined as ``Entropy Collapse Layer'', which generalizes across diverse datasets and MLLMs.
Motivated by this discovery, we apply matrix entropy to the token level to precisely quantify the information content of individual visual tokens, enabling targeted pruning at the Entropy Collapse Layer.

\section{Methodology}
\label{sec:methodology}

\subsection{Preliminaries}
\label{Preliminaries}

We first introduce matrix entropy for characterizing the information content of a visual token sequence. 
Let $\mathbf{X}=[\mathbf{x}_1,\mathbf{x}_2,\ldots,\mathbf{x}_N]$ denote the visual token matrix, where $\mathbf{x}_i\in\mathbb{R}^d$ is the $i$-th token and $N$ is the number of visual tokens. 
The trace-normalized covariance matrix $\boldsymbol{\Sigma}_{\mathbf{X}}\in\mathbb{R}^{d\times d}$ is computed as:
\begin{equation}
    \boldsymbol{\Sigma_{\mathbf{X}}} = \frac{1}{N} \sum_{i=1}^{N} \frac{(\mathbf{x}_i - \bar{\mathbf{x}})(\mathbf{x}_i - \bar{\mathbf{x}})^T}{\|\mathbf{x}_i - \bar{\mathbf{x}}\|^2},
\end{equation}
where $\bar{\mathbf{x}}=\frac{1}{N}\sum_{i=1}^{N}\mathbf{x}_i$ is the mean vector.
This formulation ensures that $\text{tr}(\boldsymbol{\Sigma}_{\mathbf{X}}) = 1$, fulfilling the requirement for defining matrix entropy.
Following~\citep{giraldo2014measures}, the order-$\alpha$ matrix entropy associated with $\boldsymbol{\Sigma}_{\mathbf{X}}$ is defined as:
\begin{equation}
S_{\alpha}(\boldsymbol{\Sigma}_{\mathbf{X}})
=\frac{1}{1-\alpha}\log\!\left(\mathrm{tr}\!\left(\boldsymbol{\Sigma}_{\mathbf{X}}^{\alpha}\right)\right).
\end{equation}

\begin{figure}[t]
\begin{center}
\centerline{\includegraphics[width=1.0\linewidth]{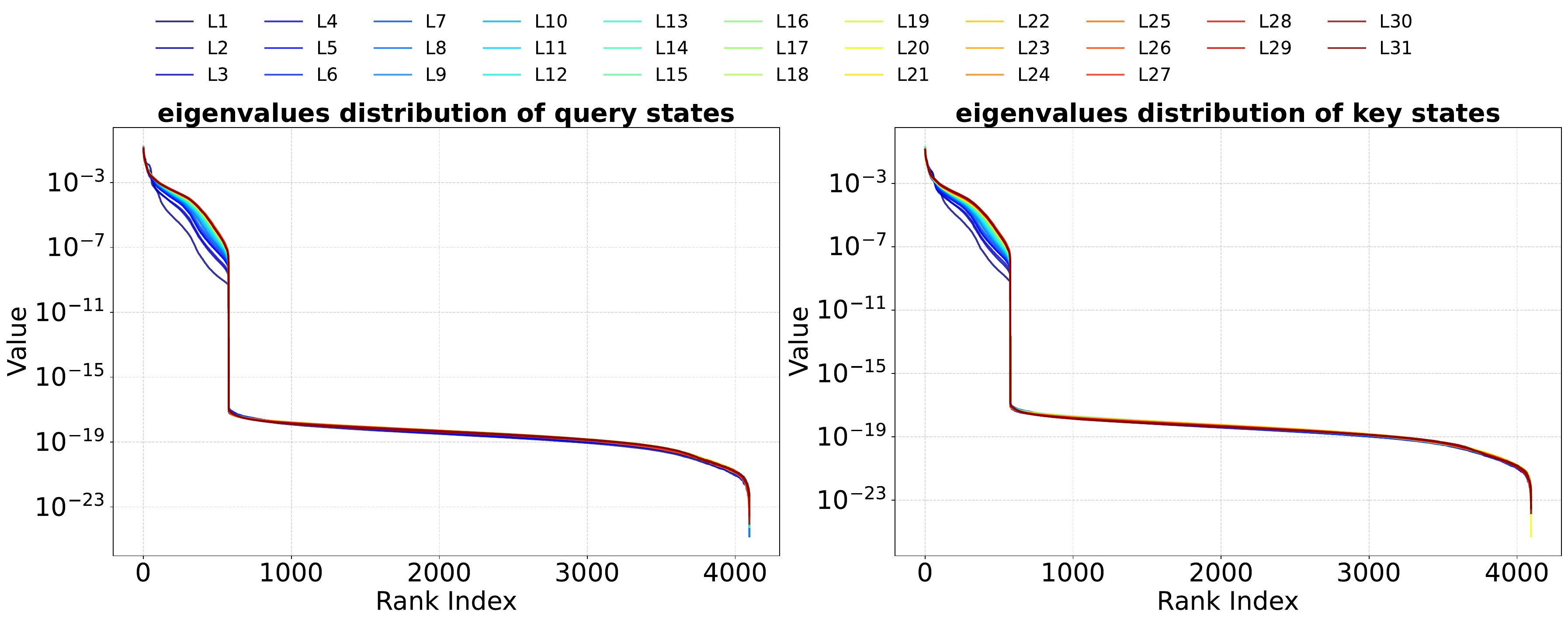}}
\caption{Eigenvalue distribution of the query and key states covariance matrices. L$i$ represents the $i-th$ layer. We visualize the magnitude of eigenvalues across different layers of LLaVA-1.5-7B. The rapid decay observed in the eigenvalues distribution indicates a low-rank structure within the matrices.
}
\vspace{-1cm}
\label{fig:eigen_distribution}
\end{center}
\end{figure}

\begin{figure*}[t]
\begin{center}
\centerline{\includegraphics[width=1.0\linewidth]{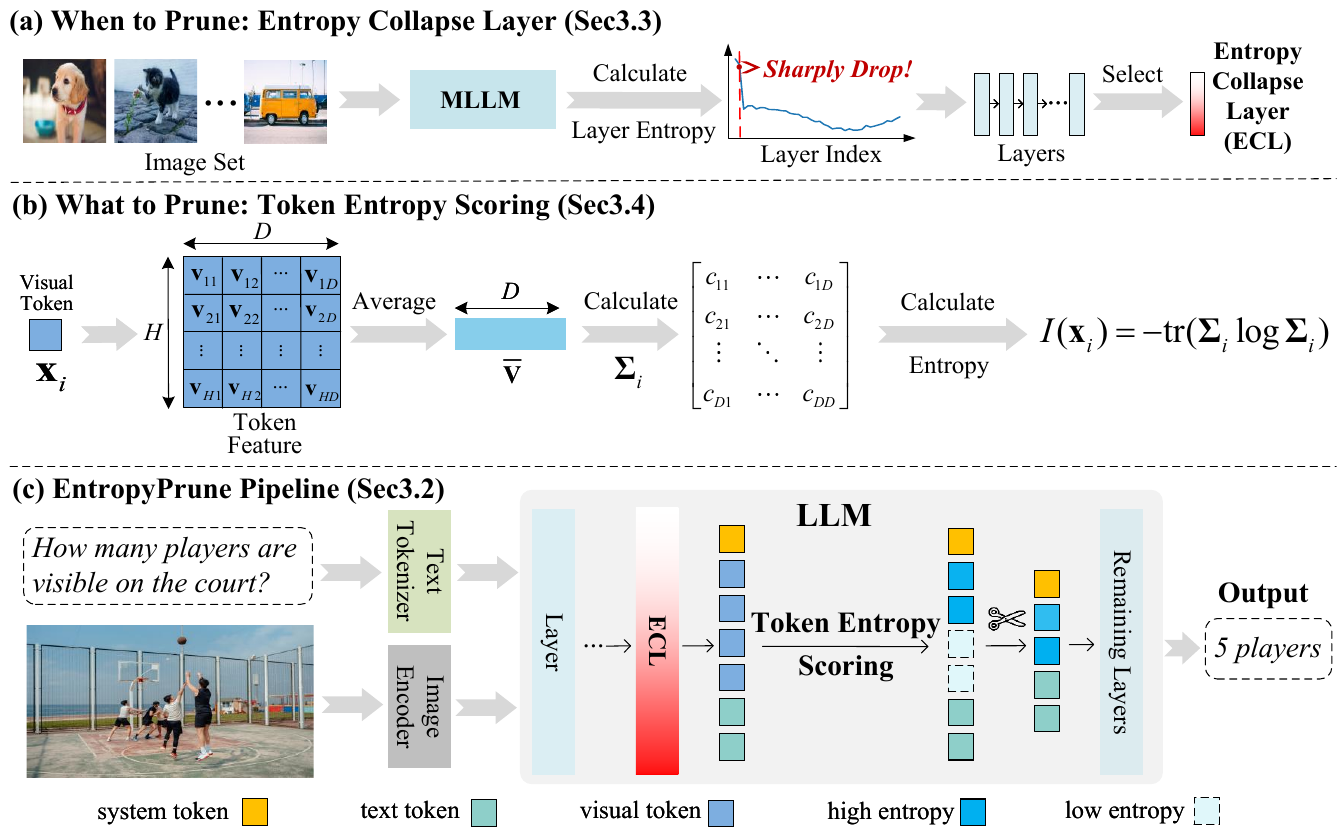}}
\vspace{-1mm}
\caption{Overview of the proposed EntropyPrune. 
(a) \textbf{When to Prune} identifies the ``Entropy Collapse Layer'' by detecting a sharp drop in layer-wise matrix entropy. 
(b) \textbf{What to Prune} details the calculation of matrix entropy based on the token's feature.
(c) \textbf{EntropyPrune Pipeline} illustrates the overall workflow: after reaching ``Entropy Collapse Layer'', the matrix entropy of each visual token is calculated to guide the pruning of low-entropy ones.
}
\vspace{-10mm}
\label{fig:framework}
\end{center}
\end{figure*}

\noindent\textbf{Lemma 1.} \textit{
Let $\{\sigma_i\}$ be the eigenvalues of $\boldsymbol{\Sigma}_{\mathbf{X}}$. Then
}
\begin{equation}
S_{\alpha}(\boldsymbol{\Sigma}_{\mathbf{X}})
=\frac{1}{1-\alpha}\log\!\left(\sum_i\sigma_i^{\alpha}\right).
\end{equation}
The proof is provided in Appendix~\ref{app:proof_lemma1}.

\noindent\textbf{Lemma 2} \textit{
As $\alpha\rightarrow 1$, the order-$1$ matrix entropy satisfies:}
\begin{equation}
    S(\boldsymbol{\Sigma_{\mathbf{X}}}) = \lim_{\alpha \to 1} S_\alpha(\boldsymbol{\Sigma_{\mathbf{X}}}) = 
    - \sum_i \sigma_i \log \sigma_i,
\end{equation}
where we adopt the convention $0 \log 0 = 0$. The proof of this lemma is provided in Appendix~\ref{app:proof_lemma2}.

\noindent\textbf{Lemma 3.} \textit{
The trace-normalized covariance matrix $\boldsymbol{\Sigma_{\mathbf{X}}}$ functions as the density matrix $\rho$ in quantum mechanics, and the order-1 matrix entropy $S(\boldsymbol{\Sigma_{\mathbf{X}}})$ is mathematically equivalent to the Von Neumann entropy defined in quantum statistical mechanics:
}
\begin{equation}
    S(\boldsymbol{\Sigma_{\mathbf{X}}}) \equiv -\mathrm{tr}(\rho \log \rho),
\end{equation}
The proof is provided in Appendix~\ref{app:proof_lemma3}.

To empirically support this connection, we visualize the eigenvalue distributions of $\boldsymbol{\Sigma_{\mathbf{X}}}$ computed from the query and key states across layers of LLaVA-1.5-7B. 
As shown in Figure~\ref{fig:eigen_distribution}, the spectra are highly concentrated, indicating that most variance (and thus information) is captured by a small number of principal components. 
Motivated by this observation, we approximate the matrix entropy using the top-$k$ eigenvalues:
\begin{equation}
S(\boldsymbol{\Sigma_{\mathbf{X}}}) \approx - \sum_{i=1}^{k} \sigma_i \log \sigma_i,
\end{equation}
where $\sigma_1 \ge \sigma_2 \ge \dots \ge \sigma_k$ are the top-$k$ eigenvalues of $\boldsymbol{\Sigma_{\mathbf{X}}}$. 
For brevity, we refer to $S(\boldsymbol{\Sigma_{\mathbf{X}}})$ as \emph{matrix entropy} in the remainder of this paper.

\subsection{EntropyPrune Framework}
\label{sec:framework}
Figure~\ref{fig:framework} illustrates the overall pipeline of EntropyPrune. 
Given an MLLM, we (i) analyze layer-wise matrix entropy to locate the \emph{Entropy Collapse Layer} (ECL), where a sharp information drop indicates increasing redundancy (Figure~\ref{fig:framework}(a)); 
(ii) quantify the information richness of individual visual tokens via matrix entropy (Figure~\ref{fig:framework}(b)); and 
(iii) prune low-entropy tokens and forward compact representations for efficient generation (Figure~\ref{fig:framework}(c)).

\subsection{When to Prune: Entropy Collapse Layer}
\label{sec:ecl}
As shown in Figure~\ref{fig:layer_matrix_entropy_llava}, matrix entropy follows a highly consistent depth-wise pattern across eight datasets, despite substantial variation in input images and instructions. 
This suggests that matrix entropy captures an intrinsic redundancy trend of multimodal representations, robust to variations in input data. 
While prior work reports a general information decay with depth~\citep{xing2024pyramiddrop,lin2025boosting,wang2026tokenpruningworserandom}, we observe a critical phenomenon within this trend.
Specifically, the matrix entropy of query and key states remains high in early layers but drops abruptly after a certain depth (\textit{e.g.}, the second layer for LLaVA-1.5-7B and LLaVA-NeXT-7B). 
We refer to this transition point as the \textbf{Entropy Collapse Layer} (ECL). 
The collapse indicates rapid compression of visual evidence, after which many tokens become informationally dispensable. 
Accordingly, ECL serves as an interpretable criterion for selecting the pruning stage, avoiding heuristic layer choices.

\subsection{What to Prune: Token Entropy Scoring}
\label{sec:token_scoring}
Given the pruning stage (\textit{i.e.}, ECL), we rank visual tokens by their information content using token-wise matrix entropy. 
The feature matrix $\mathbf{M}_i\in\mathbb{R}^{h\times d_h}$ for each token $\mathbf{x}_i$ is given by:
\begin{equation}
\mathbf{M}_i=[\mathbf{v}_1,\mathbf{v}_2,\dots,\mathbf{v}_h]^{\top},
\end{equation}
where $\mathbf{v}_j\in\mathbb{R}^{d_h}$ denotes the feature for the $j$-th attention head and 
$\bar{\mathbf{v}}=\frac{1}{h}\sum_{j=1}^{h}\mathbf{v}_j$ is the mean feature. 
Since independent attention heads focus on diverse patterns, constructing $\mathbf{M}_i$ in this manner allows us to evaluate the richness of token information across different representational views.
The trace-normalized covariance matrix of the token $\mathbf{x}_i$ is calculated as:
\begin{equation}
\boldsymbol{\Sigma}_i=\frac{1}{h}\sum_{j=1}^{h}\frac{(\mathbf{v}_j-\bar{\mathbf{v}})(\mathbf{v}_j-\bar{\mathbf{v}})^{\top}}{\|\mathbf{v}_j-\bar{\mathbf{v}}\|^2},
\label{eq:covariance}
\end{equation}
where $\boldsymbol{\Sigma}_i \in \mathbb{R}^{d_h \times d_h}$ captures the intra-token correlation structure.

Let $\{\sigma_t\}$ be eigenvalues of $\boldsymbol{\Sigma}_i$. The token score is:
\begin{equation}
I(\mathbf{x}_i)=-\mathrm{tr}(\boldsymbol{\Sigma}_i\log\boldsymbol{\Sigma}_i)
=-\sum_{t=1}^{d_h}\sigma_t\log\sigma_t.
\label{eq:token_entropy}
\end{equation}
Higher scores indicate more diverse information distribution, whereas lower scores suggest redundancy. 
EntropyPrune retains high-score tokens and prunes the rest.

\subsection{How to Compute Fast: Spectral Acceleration}
\label{sec:acceleration}
Direct eigendecomposition of $\boldsymbol{\Sigma}_i\in\mathbb{R}^{d_h\times d_h}$ costs $\mathcal{O}(d_h^3)$ time and is expensive in practice. 
In typical MLLM architectures (\textit{e.g.}, LLaVA-1.5, Qwen2.5-VL), the head dimension $d_h$ is often much larger than the number of heads $h$. For instance, with $d_h=128$ and $h=32$ in LLaVA-1.5-7B, the computational cost becomes prohibitive for real-time inference.
To address this issue, we propose a \textbf{Spectral Acceleration Strategy} that exploits the identical non-zero spectrum property of dual matrices.  
Let $\tilde{\mathbf{M}}_i \in \mathbb{R}^{h \times d_h}$ denote the centered matrix of $\mathbf{x}_i$:
\begin{equation}
    \tilde{\mathbf{M}}_i = \mathbf{M}_i - \mathbf{1}\bar{\mathbf{v}}^T = [\mathbf{v}_1 - \bar{\mathbf{v}}, \dots, \mathbf{v}_h - \bar{\mathbf{v}}]^T.
\end{equation}
To ensure that the resulting covariance matrix is strictly trace-normalized, we perform $L_2$ normalization on each row of the centered matrix:
\begin{equation}
    \tilde{\mathbf{M}}_i \leftarrow \left[ \frac{\mathbf{v}_1 - \bar{\mathbf{v}}}{\|\mathbf{v}_1 - \bar{\mathbf{v}}\|_2}, \dots, \frac{\mathbf{v}_h - \bar{\mathbf{v}}}{\|\mathbf{v}_h - \bar{\mathbf{v}}\|_2} \right]^T.
\end{equation}
Accordingly, the original traced-normalized covariance matrix can be rewritten as the Gram matrix of the columns of $\tilde{\mathbf{M}}_i$:
\begin{equation}
    \boldsymbol{\Sigma}_i = \frac{1}{h} \tilde{\mathbf{M}}_i^T \tilde{\mathbf{M}}_i \in \mathbb{R}^{d_h \times d_h}.
\end{equation}
We define its dual counterpart, the Gram matrix of $\boldsymbol{\Sigma}_i$ as $\tilde{\boldsymbol{\Sigma}}_i$:
\begin{equation}
    \tilde{\boldsymbol{\Sigma}}_i = \frac{1}{h} \tilde{\mathbf{M}}_i \tilde{\mathbf{M}}_i^T  \in \mathbb{R}^{h \times h}.
\end{equation}

Since $A^{\top}A$ and $AA^{\top}$ share identical non-zero eigenvalues, $\boldsymbol{\Sigma}_i$ and $\tilde{\boldsymbol{\Sigma}}_i$ have the same spectrum. 
Therefore, we compute the exact entropy using $\tilde{\boldsymbol{\Sigma}}_i$:
\begin{equation}
I(\mathbf{x}_i)
=-\mathrm{tr}(\tilde{\boldsymbol{\Sigma}}_i\log\tilde{\boldsymbol{\Sigma}}_i)
=-\sum_{t=1}^{h}\sigma_t\log\sigma_t,
\end{equation}
reducing the complexity to $\mathcal{O}(h^3)$. 
For typical settings ($d_h=128$, $h=32$), this yields a $64\times$ theoretical speedup.

\renewcommand{\multirowsetup}{\centering}
\definecolor{mygray}{gray}{.92}
\definecolor{ForestGreen}{RGB}{34,139,34}
\newcommand{\fg}[1]{\mathbf{\mathcolor{ForestGreen}{#1}}}
\definecolor{Forestred}{RGB}{220,50,50}
\newcommand{\fr}[1]{\mathbf{\mathcolor{Forestred}{#1}}}
\begin{table*}[t]
    \centering
    \setlength{\tabcolsep}{4.5pt}
    \renewcommand{\arraystretch}{1.33}
    \footnotesize
	\centering
    \caption{\textbf{Performance of LLaVA-1.5-7B with EntropyPrune.} The vanilla number of vision tokens is 576. \textbf{Acc.} represents the average accuracy across all benchmarks. \textbf{Rel.} denotes the relative performance retained compared to the original model. The best performance is highlighted in \textbf{bold}.}
    \vspace{-1mm}
	\label{tab:llava15}
    \begin{tabular}{p{3.0cm}|c c c c c c c c}
        \shline
        \textbf{Method} & \textbf{MMB}  & \textbf{MME} & \textbf{VQA}$^{\text{T}}$ & \textbf{MMB}$^{\text{C}}$ & \textbf{MMstar} & \textbf{Acc.(\%)} & \textbf{Rel.(\%)} & \textbf{FLOPs} \\
        \shline
        \rowcolor{mygray}
        \multicolumn{9}{c}{\textit{Upper Bound, 576 Tokens} \ $\textbf{(100\%)}$}\\
        LLaVA-1.5-7B & 64.6  & 1862 & 58.3 & 58.1 & 33.1 & 56.1 & 100.0 & 8.80T \\ 
        \hline

        \rowcolor{mygray}
        \multicolumn{9}{c}{\textit{Retain 192 Tokens} \ $\fg{(\downarrow 66.7\%)}$} \\
        FastV \texttt{\scriptsize{(ECCV24)}} & 63.2  & 1796 & 56.4 & 57.4 & 31.1 & 54.4 & 97.0 & 3.99T \\ 
        PDrop \texttt{\scriptsize{(CVPR25)}} & 63.3  & 1765 & 56.2 & 56.5 & 30.7 & 53.9 & 96.1 & 3.79T \\
        DART \texttt{\scriptsize{(EMNLP25)}}& 63.7 & 1833 & 57.0  & 57.1  & 31.4 & 54.9 & 97.9 & 3.69T \\
        CDPruner \texttt{\scriptsize{(NIPS25)}}& 63.2 & 1786 & 57.3  & 55.6 & 31.1 & 54.2 & 96.6 & 3.76T \\
        DivPrune \texttt{\scriptsize{(CVPR25)}} & 62.3  & 1769 & 56.6 & 56.2 & 31.5 & 54.0 & 96.1 & 3.73T \\
        Prumerge \texttt{\scriptsize{(ICCV25)}} & 56.3  & 1515 & 50.1 & 51.3 & 22.0 & 46.8 & 83.3 & 3.71T \\
        EntropyPrune & \textbf{64.4} & \textbf{1844} & \textbf{57.6} & \textbf{57.9} & \textbf{31.5} & \textbf{55.5} & \textbf{98.8} & 3.72T \\

        \shline
	\end{tabular}
    \vspace{-6mm}
\end{table*}

\subsection{Theoretical Analysis of Computational Complexity}
\label{sec:analysis_computaion}
In this analysis, we exclusively focus on the LLM backbone of MLLMs, taking LLaVA-1.5-7B as a representative example. As text tokens are typically much shorter than visual tokens, we concentrate on the FLOPs contributed by visual tokens. Let $n$ denote the number of visual tokens, $d$ the hidden size, and $m$ the FFN intermediate size (with SwiGLU), where typically $m \approx \frac{8}{3}d$. For the prefill stage, the FLOPs per transformer layer can be approximated as:
\begin{equation}
\begin{split}
F(n) &= 4n^2d + 8nd^2 + 6nmd \\
&\approx 4n^2d + 24nd^2.
\end{split}
\end{equation}
For the detailed derivation, please refer to the Appendix~\ref{app:derive_flops_llava}.
Suppose the total number of LLM layers is $L$ and the token pruning is applied at the $k$-th layer.
If the token count is reduced by a ratio $r$ ($\hat{n} = (1-r)n$), the total FLOPs reduction ratio is calculated as:
\begin{equation}
R = 1 - \frac{k \cdot F(n) + (L-k)\cdot F(\hat{n})}{L \cdot F(n)}.
\end{equation}
Given $d \gg n$ in LLaVA-1.5 configurations (\textit{e.g.}, $d=4096$ vs $n = 576$), the linear term $24nd^2$ dominates the computation over the quadratic attention term $4n^2d$. Thus, $\frac{F(\hat{n})}{F(n)} \approx 1 - r$, and the overall reduction ratio simplifies to $R \approx \frac{L-k}{L} r$.

Furthermore, the additional computational overhead introduced by EntropyPrune is negligible compared to the backbone inference.
Specifically, the computational costs for computing the covariance matrix $\tilde{\boldsymbol{\Sigma}}_i$ and performing the eigenvalue decomposition are $2nhd$ and $4nh^3$, respectively~\citep{matrixcomputation}.
Consequently, the FLOPs of EntropyPrune is approximately $96nd$. 
Comparing this to a single Transformer layer, the overhead ratio is roughly $\frac{96nd}{4n^2d + 24nd^2} \approx \frac{4}{d}$. 
Given the typically large hidden dimension $d$, this fraction is vanishingly small, rendering the overhead of EntropyPrune practically negligible.

\section{Experiment}
\label{sec:experiment}

\subsection{Experiment Setup}
\label{Exp_stt}


\textbf{Base Models.}
We adopt the LLaVA series to evaluate our method across different modalities: LLaVA-1.5~\citep{liu2024improved} (general image), LLaVA-NeXT~\citep{liu2024llavanext} (high-resolution image), and Video-LLaVA~\citep{2024videollava} (video). Additionally, we incorporate Qwen2.5-VL~\citep{bai2025qwen2} to examine performance on cutting-edge open-source architectures.

\noindent\textbf{Baselines.}
We compare our approach with several representative MLLMs token pruning methods, including  FastV~\citep{chen2024image}, 
PDrop~\citep{xing2024pyramiddrop}, 
DART~\citep{wen2025stop},
DivPrune ~\citep{alvar2025divprune},
CDPruner ~\citep{zhang2026beyond},
LLaVA-PruMerge(Prumerge)~\citep{shang2025prumerge}.
More details are provided in Appendix~\ref{app:baseline}.


\renewcommand{\multirowsetup}{\centering}
\definecolor{mygray}{gray}{.92}
\definecolor{ForestGreen}{RGB}{34,139,34}
\definecolor{Forestred}{RGB}{220,50,50}
\begin{table}[t]
    \centering
    \setlength{\tabcolsep}{1pt}
    \renewcommand{\arraystretch}{1.33}
    \footnotesize
	\centering
    \caption{\textbf{Performance of LLaVA-Next-7B with EntropyPrune.} The best performance is highlighted in \textbf{bold}, while the second best is \underline{underlined}.}
    \vspace{-3mm}
	\label{tab:llavanext}
    \begin{tabular}{p{2cm}|c c c c c c}
        \shline
        \textbf{Method} & \textbf{VQA}$^{\text{T}}$ & \textbf{MMVet}  & \textbf{OCR$^{\text{B}}$} & \textbf{MMMU} & \textbf{AI2D} & \textbf{Acc.} \\
        \shline
        \rowcolor{mygray}
        \multicolumn{7}{c}{\textit{Upper Bound, 2880 Tokens} \ $\textbf{(100\%)}$}\\
        LLaVA-Next & 61.3 & 44.0 & 43.7 & 28.0 & 65.6  & 48.5 \\

        \hline

        \rowcolor{mygray}
        \multicolumn{7}{c}{\textit{Retain 320 Tokens} \ $\fg{(\downarrow 88.9\%)}$} \\
        FastV & 49.0 & 31.1 & 21.8 & \underline{29.3} & 60.2 & 38.3 \\ 
        PDrop & 53.2 & 29.4 & 27.1  & 27.3 & 61.5 & 39.7 \\
        DART & \underline{58.0} & \underline{39.3} & \underline{30.5} & 26.7 & \underline{62.2} & 43.3 \\
        CDPruner & 57.4 & 38.0 & \textbf{33.1} & 27.3 & 61.6 & \underline{43.5} \\
        DivPrune & 51.4 & 34.4 & 29.4 & 25.3 & 60.9 & 40.3 \\
        EntropyPrune & \textbf{58.5} & \textbf{43.8} & \underline{30.5} & \textbf{30.0} & \textbf{62.5} & \textbf{45.1} \\
        \hline
        
        \shline
	\end{tabular}
    \vspace{-6mm}
\end{table}

\subsection{Performance on general token pruning}
\label{sec:llava15}
As shown in~\cref{tab:llava15}, on the LLaVA-1.5-7B, EntropyPrune consistently surpasses all competing baselines when retaining only 192 visual tokens. 
For instance, EntropyPrune remarkably achieves 55.5\% average accuracy and maintains 98.8\% relative performance while removing 66.7\% of tokens.
Notably, EntropyPrune closely approaches the upper bound performance on benchmarks like MMB and MME while retaining only 192 visual tokens, indicating that our method effectively identifies low-information visual tokens that may otherwise hinder model performance.
Overall, EntropyPrune reduces FLOPs by 57.7\% , with only a minimal average performance drop of 0.6\%.

\subsection{Performance on high resolution inputs}
\label{sec:llavanext}

To further demonstrate the effectiveness of EntropyPrune when taking high-resolution images as input, we evaluate it on the LLaVA-NeXT-7B, which is capable of processing high resolution images.
As shown in~\cref{tab:llavanext}, EntropyPrune demonstrates remarkable scalability, achieving 45.1\% average accuracy across five benchmarks while retaining only 11.1\% of visual tokens. 
Notably, EntropyPrune surpasses the base model by 2.0\% on MMMU and outperforms the second-best baseline, DART, by 4.5\% on MM-Vet. These results underscore the efficacy of EntropyPrune in processing high-resolution inputs.

\begin{table}[t]
    \centering
    \setlength{\tabcolsep}{1pt}
    \renewcommand{\arraystretch}{1.33}
    \footnotesize
	\centering
	\caption{\textbf{Performance of EntropyPrune on Qwen-2.5-VL-7B.}}
    \vspace{-3mm}
	\label{tab:qwen25}
    \begin{tabular}{p{2cm}|c c c c c c }
        \shline
        \textbf{Method} & \textbf{MMB} & \textbf{MMB}$^{\text{C}}$  & \textbf{MMstar} & \textbf{MMMU} & \textbf{AI2D} & \textbf{Acc.} \\
        \shline
        \rowcolor{mygray}
        \multicolumn{7}{c}{\textit{Upper Bound} \ $\textbf{(100\%)}$}\\
        Qwen2.5-VL & 79.8 & 81.0 & 56.3 & 58.0 & 80.8  & 71.2  \\ 
        \hline

        \rowcolor{mygray}
        \multicolumn{7}{c}{\textit{Retain 25\% Tokens} \ $\fg{(\downarrow 75\%)}$} \\
        FastV & 77.7 & 77.2 & 46.7 & 42.7 & \textbf{74.5} & 63.8 \\ 
        CDPruner & 74.3 & 77.7 & 47.8 & 42.9 & 71.6 & 62.9 \\ 
        EntropyPrune & \textbf{77.9} & \textbf{79.6} & \textbf{50.6} & \textbf{52.1} & 74.1 & \textbf{66.9} \\
        \hline

        \rowcolor{mygray}
        \multicolumn{7}{c}{\textit{Retain 12.5\% Tokens} \ $\fg{(\downarrow 87.5\%)}$} \\
        FastV  & 53.6 & 60.0 & 31.9 & 38.2 & 65.5 & 49.8  \\ 
        CDPruner & 71.6 & 74.7 & 43.7 & 42.1 & 68.3 & 60.1 \\ 
        EntropyPrune & \textbf{73.3} & \textbf{75.8} & \textbf{44.5} & \textbf{48.8} & \textbf{69.4} & \textbf{62.4} \\
        
        \shline
	\end{tabular}
    \vspace{-7mm}
\end{table}

\subsection{EntropyPrune with Qwen architecture}
\label{sec:qwen25}

Qwen2.5-VL series adopts a more advanced MLLM architecture that supports Naive Dynamic Resolution to process images of arbitrary aspect ratios.
To verify the architectural robustness of EntropyPrune, we conduct experiments on the Qwen2.5-VL-7B.
As presented in~\cref{tab:qwen25}, EntropyPrune consistently surpasses competing baselines across different pruning ratios.
When retaining 25\% of visual tokens, EntropyPrune achieves an average accuracy of 66.9\%, outperforming FastV and CDPruner by 3.1\% and 4.0\%, respectively.
Even under aggressive pruning where only 12.5\% of tokens are retained, EntropyPrune maintains a robust average accuracy of 62.4\%, significantly exceeding FastV~(49.8\%) and CDPruner~(60.1\%).
Notably, on the challenging MMMU benchmark, EntropyPrune preserves significantly better performance than CDPruner (48.8\% vs 42.1\%), demonstrating the robust performance across diverse model architectures.

\begin{table}[t]
    \centering
    \setlength{\tabcolsep}{4pt}
    \renewcommand{\arraystretch}{1.33}
    \footnotesize
	\centering
	\caption{\textbf{Performance of EntropyPrune on Video-LLaVA-7B.}}
    \vspace{-1mm}
	\label{tab:videollava}
    \begin{tabular}{p{1.85cm}|c c c c c c}
        \shline
        \multirow{2}{*}{\textbf{Method}} & \multicolumn{2}{c}{\textbf{MSVD}} & \multicolumn{2}{c}{\textbf{MSRVTT}}  & \multicolumn{2}{c}{\textbf{Avg.}}  \\
        & Acc. & Score & Acc. & Score & Acc. & Score \\ 
        \shline
        \rowcolor{mygray}
        \multicolumn{7}{c}{\textit{Upper Bound} \ $\textbf{(100\%)}$}\\
        Video-LLaVA & 53.1 & 3.2 & 35.6 & 2.5  & 44.4 & 2.9 \\
        \hline

        \rowcolor{mygray}
        \multicolumn{7}{c}{\textit{Retain 50\% Tokens} \ $\fg{(\downarrow 50\%)}$} \\ 
        CDPruner &  41.9 & 2.6 & 30.1 & 2.2 & 36.0 & 2.4 \\ 
        EntropyPrune & \textbf{52.8} & \textbf{3.2} & \textbf{36.0} & \textbf{2.5} & \textbf{44.4} & \textbf{2.9} \\
        \shline
	\end{tabular}
    \vspace{-5mm}
\end{table}

\subsection{EntropyPrune with Video tasks}
\label{sec:videollava}

To verify the capability of EntropyPrune in handling video datas, we integrate it with Video-LLaVA-7B and conduct experiments on the MSVD-QA(MSVD) and MSRVTT-QA(MSRVTT) benchmarks.
As presented in~\cref{tab:videollava}, EntropyPrune consistently surpasses CDPruner when retaining 50\% of visual tokens.
Specifically, it achieves an average accuracy of 44.4\%, outperforming CDPruner by a significant margin of 8.4\%.
Notably, on the MSRVTT benchmark, EntropyPrune achieves 36.0\% accuracy, which slightly outperforms the base model (35.6\%).
These results demonstrates that EntropyPrune effectively eliminates redundant spatiotemporal tokens while preserving the essential visual cues necessary for video reasoning.

\subsection{Efficiency Comparison.}
\label{sec:efficiency}

To evaluate the efficiency of EntropyPrune, we conduct a comparative analysis against FastV and CDPruner on the LLaVA-1.5-7B. We report prefilling time (Prefill), latency (Lat.), KV cache (KV), GPU memory (Mem.), and FLOPs. The MME benchmark is selected for this evaluation as it encompasses one prefill and one decode stage. As reported in~\cref{tab:eff}, when retaining 128 tokens, EntropyPrune achieves 1.6$\times$ and 1.4$\times$ acceleration in prefilling time and latency, respectively, with only a marginal 4.6\% decrease in total score. At this retention level, EntropyPrune reduces KV cache by 77.8\% and FLOPs by 67.3\%. Compared to FastV and CDPruner, EntropyPrune demonstrates superior efficiency by consuming less GPU memory and achieving faster inference speeds, while maintaining better performance.

\subsection{Ablation Study}
\label{sec:ablation}
In this section, we perform ablation studies to validate the two core contributions of our work: the effectiveness of \textbf{Entropy Collapse Layer} and our token entropy scoring strategy. All experiments are conducted on the LLaVA-1.5-7B.

\renewcommand{\multirowsetup}{\centering}
\definecolor{mygray}{gray}{.92}
\definecolor{ForestGreen}{RGB}{34,139,34}
\definecolor{Forestred}{RGB}{220,50,50}
\begin{table}[t]
    \centering
    \setlength{\tabcolsep}{2pt}
    \renewcommand{\arraystretch}{1.33}
    \footnotesize
	\centering
    \caption{\textbf{Efficiency analysis of different pruning methods on LLaVA-1.5-7B.}}
    \vspace{-1mm}
	\label{tab:eff}
    \begin{tabular}{p{1.95cm}|c c c c c c}
        \shline
        \textbf{Method} &  \begin{tabular}[c]{@{}c@{}}\textbf{Prefill}\\ \textbf{(s)}\end{tabular} & \begin{tabular}[c]{@{}c@{}}\textbf{Lat.}\\ \textbf{(s)}\end{tabular} & \begin{tabular}[c]{@{}c@{}}\textbf{KV}\\ \textbf{(MB)}\end{tabular} & \begin{tabular}[c]{@{}c@{}}\textbf{Mem.}\\ \textbf{(GB)}\end{tabular} & \begin{tabular}[c]{@{}c@{}}\textbf{FLOPs}\\ \textbf{(\%)}\end{tabular} & \textbf{Score} \\
        \shline
        \rowcolor{mygray}
        \multicolumn{7}{c}{\textit{Upper Bound, 576 Tokens} \ $\textbf{(100\%)}$}\\
        LLaVA-1.5-7B & 381.8 & 459.6 & 288.0 & 16.2  & 100 & 1862   \\ 
        \hline

        \rowcolor{mygray}
        \multicolumn{7}{c}{\textit{Retain 192 Tokens} \ $\fg{(\downarrow 66.7\%)}$} \\
        FastV  & 297.5 & 370.3 & 96.3 & 16.0 & 45.3 & 1796  \\ 
        CDPruner & 360.4 & 415.7 & 96.0 & 19.8  & 42.7 & 1786  \\
        EntropyPrune & \textbf{273.3} & \textbf{358.2} & \textbf{95.8} & \textbf{15.8} & \textbf{42.3} & \textbf{1844}  \\
        \hline

        \rowcolor{mygray}
        \multicolumn{7}{c}{\textit{Retain 128 Tokens} \ $\fg{(\downarrow 77.8\%)}$}\\
        FastV  & 256.2 & 341.1 & 64.2 & 15.9 & 35.7 & 1735  \\ 
        CDPruner & 294.2 & 352.0 & 64.0 & 19.7 & 33.2  & 1775  \\
        EntropyPrune & \textbf{244.3} & \textbf{330.1} & \textbf{63.9} & \textbf{15.7} & \textbf{32.7} & \textbf{1780}   \\
        
        \shline
	\end{tabular}
    \vspace{-3mm}
\end{table}

\begin{figure}[t]
\begin{center}
\centerline{\includegraphics[width=1.0\linewidth]{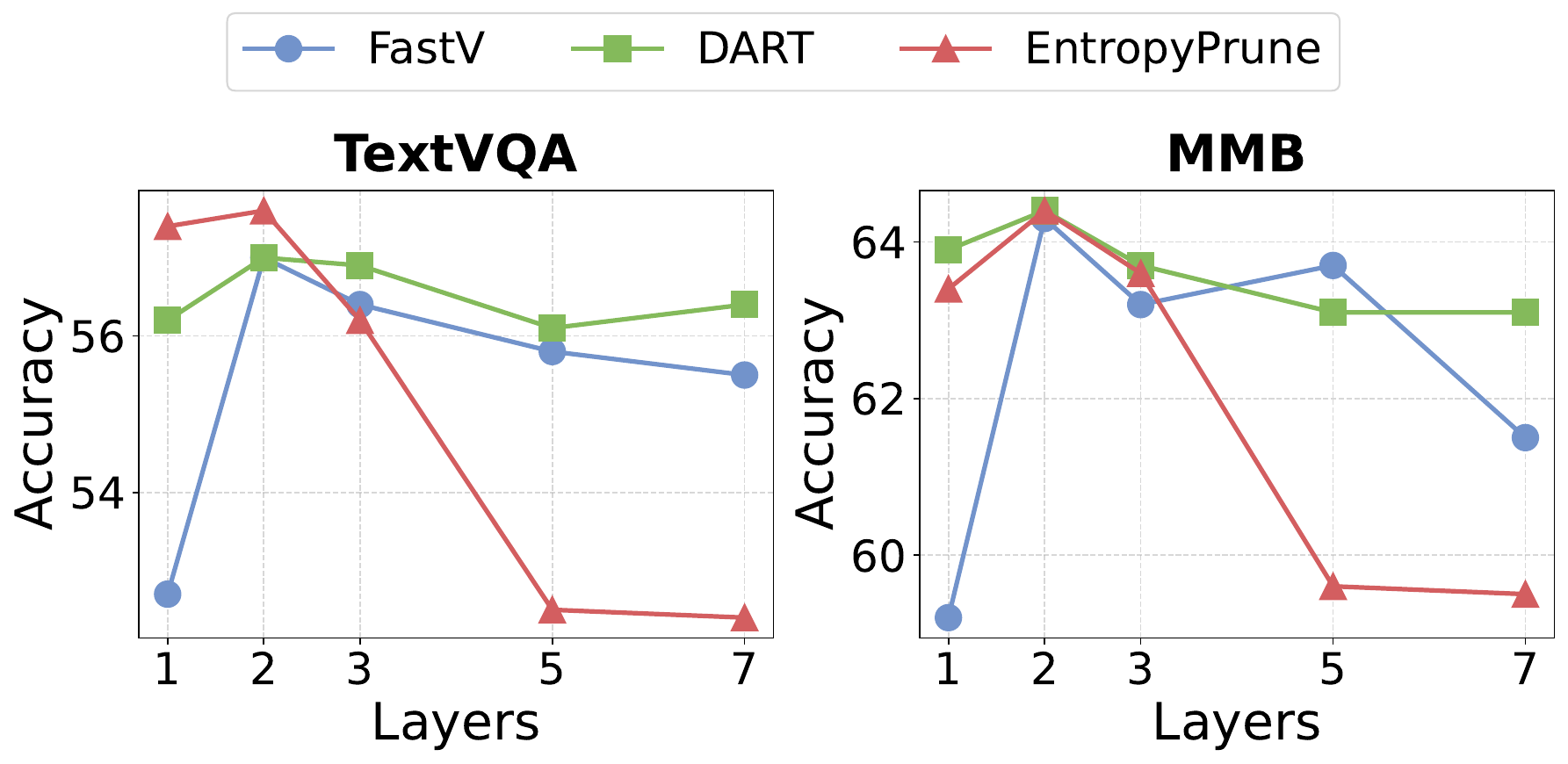}}
\vspace{-2mm}
\caption{
\textbf{Ablation study on pruning layer selection.} 
Experiments are conducted on TextVQA and MMB when retaining 192 tokens. Applying pruning at \textbf{Entropy Collapse Layer (Layer 2)} consistently yields the best performance across all methods.
}
\label{fig:ablation_layer}
\vspace{-11mm}
\end{center}
\end{figure}

\renewcommand{\multirowsetup}{\centering}
\definecolor{mygray}{gray}{.92}
\definecolor{ForestGreen}{RGB}{34,139,34}
\definecolor{Forestred}{RGB}{220,50,50}
\begin{table}[t]
    \centering
    \setlength{\tabcolsep}{3pt}
    \renewcommand{\arraystretch}{1.33}
    \footnotesize
	\centering
    \caption{\textbf{Ablation study on token selection strategies.}}
    \vspace{-1mm}
	\label{tab:ablation_method}
    \begin{tabular}{p{2.2cm}|c c c c c}
        \shline
        \textbf{Method} & \textbf{MMB}  & \textbf{MME} & \textbf{SQA}  & \textbf{VQA}$^{\text{T}}$ & \textbf{Rel.(\%)} \\
        \shline
        \rowcolor{mygray}
        \multicolumn{6}{c}{\textit{Upper Bound, 576 Tokens} \ $\textbf{(100\%)}$}\\
        LLaVA-1.5-7B & 64.6  & 1862 & 69.5 & 58.3 & 100.0 \\ 
        \hline

        \rowcolor{mygray}
        \multicolumn{6}{c}{\textit{Retain 192 Tokens} \ $\fg{(\downarrow 66.7\%)}$} \\
        FastV  & 64.3  & 1821 & 68.9 & 57.0 & 98.6 \\ 
        DART & \textbf{64.4} & 1833 & 69.0 & 57.0  & 98.8 \\
        DivPrune  & 63.2  & 1795 & 68.5 & 57.5 & 97.9 \\
        EntropyPrune & \textbf{64.4} & \textbf{1844} & \textbf{69.1} & \textbf{57.6} & \textbf{99.2} \\
        \shline
	\end{tabular}
    \vspace{-5mm}
\end{table}

\subsubsection{Analysis of Pruning Layer Selection}
\label{subsec:ablation_layer}
Our analysis about layer-wise entropy posits that information of visual tokens undergoes a precipitous drop at the \textbf{Entropy Collapse Layer} (Layer 2 for LLaVA-1.5-7B), identifying it as the optimal stage for token pruning. To empirically verify this hypothesis, we evaluate the performance of EntropyPrune when applied at varying layers, specifically $\{1, 2, 3, 5, 7\}$, while maintaining a consistent budget of retaining 192 tokens on average. For a comprehensive analysis, we further extend this evaluation to FastV and DART, examining whether they exhibit similar layer-sensitivity.

As illustrated in~\cref{fig:ablation_layer}, pruning at Layer 2 consistently yields superior performance across all methods compared to earlier or deeper layers.
Specifically, EntropyPrune achieves its peak accuracy at Layer 2, achieving 57.6\% accuracy on TextVQA and 64.4\% accuracy on MMB. 
It outperforms pruning at Layer 1 and significantly surpasses pruning at deeper layers (e.g., Layer 5), where the accuracy drops sharply to 59.6\% on TextVQA and 52.5\% on MMB.
Similarly, both FastV and DART exhibit their optimal performance at Layer 2.
These results strongly validate that the Entropy Collapse Layer serves as both the theoretical and empirical ``sweet spot'' for visual token pruning.

\subsubsection{Effectiveness of Token Selection Strategy}
\label{subsec:ablation_criteria}
Building upon the identification of Entropy Collapse Layer as the optimal pruning layer, we evaluate the efficacy of our proposed token selection strategy. We evaluate our method against three baselines: FastV, DART, and DivPrune. For a fair comparison, all methods are configured to prune tokens at the Entropy Collapse Layer (Layer 2), retaining a fixed budget of 192 tokens.

\cref{tab:ablation_method} presents the comparative results across multiple benchmarks. EntropyPrune consistently outperforms all baselines, achieving the highest relative performance retention of \textbf{99.2\%}. Notably, it achieves the best scores on MME, SQA, and VQA$^{\text{T}}$, while tying for the top performance on MMB. This highlights that prioritizing high-entropy visual tokens is vital for maintaining MLLM performance despite significant token reduction.

\section{Conclusion}
\label{sec:conclusion}

In this paper, we address the significant computational inefficiency of Multimodal Large Language Models (MLLMs) caused by visual token redundancy. Unlike existing pruning approaches that rely on empirical heuristics to select pruning layer, we introduce a rigorous theoretical analysis based on Matrix Entropy. 
Our analysis reveals the ``Entropy Collapse Layer'', where visual information drops precipitously, providing a theoretical boundary for determining the optimal pruning layer.
Building on this insight, we propose EntropyPrune, a training-free method that selectively prunes tokens based on their matrix entropy. 
To ensure efficiency, we incorporate a spectral acceleration strategy using dual Gram matrices, achieving a 64$\times$ theoretical speedup in entropy computation.
Extensive evaluations demonstrate that EntropyPrune significantly reduces inference FLOPs with minimal degradation in model performance, offering a robust solution for efficient and lightweight MLLM deployment.

\section*{Limitations}

While our EntropyPrune offers significant inference acceleration for MLLMs, its reliance on white-box architecture access presents a primary constraint. Specifically, calculating the trace-normalized covariance matrices requires direct extraction of intermediate query or key states. Consequently, this methodology is intrinsically incompatible with closed-weight systems, such as the GPT and Claude families, where internal hidden states remain inaccessible via standard APIs.

\section*{Ethical considerations}

This work introduces EntropyPrune, a training-free token pruning framework designed to optimize the inference efficiency of Multimodal Large Language Models (MLLMs). Our method substantially decreases the computational overhead, carbon footprint, and energy demands typically associated with MLLM deployment. Crucially, because EntropyPrune is entirely training-free, it completely bypasses the massive energy and hardware resource consumption required for model retraining or fine-tuning. Consequently, this approach directly aligns with the objectives of Green AI, offering a highly resource-efficient and environmentally sustainable solution for artificial intelligence application.

\section*{Acknowledgement}
This research was supported in part by the Yeqisun Joint Funds of the National Natural Science Foundation of China under Grant U2441252, in part by the National Natural Science Foundation of China under Grants 62571372 and 62271155, and in part by the Computational Biology Program (25JS2840100) of Science and Technology Commission of Shanghai Municipality (STCSM).


\bibliography{custom}

\clearpage

\appendix

\section*{Appendix}

\section{Appendix for Proofs}
\subsection{Proof of Lemma1}
\label{app:proof_lemma1}

The order-$\alpha$ matrix entropy is defined as:

\begin{equation}
    S_\alpha(\boldsymbol{\Sigma_{\mathbf{X}}}) 
    = \frac{1}{1-\alpha} \log \left( \mathrm{tr}((\boldsymbol{\Sigma_{\mathbf{X}}})^\alpha) \right).
\end{equation}

Since $\boldsymbol{\Sigma_{\mathbf{X}}}$ is the trace-normalized covariance matrix of token matrix $\mathbf{X}$, it is a real symmetric matrix and thus diagonalizable by a unitary matrix $\mathbf{U}$:

\begin{equation}
\boldsymbol{\Sigma_{\mathbf{X}}} = \mathbf{U} \mathbf{\Lambda} \mathbf{U}^\dagger,
\end{equation}

where $\mathbf{\Lambda} = \text{diag}(\sigma_1, \sigma_2, \ldots)$ is the diagonal matrix of the eigenvalues $\sigma_i$, $\mathbf{U}$ is the unitary matrix whose columns are the orthonormal eigenvectors of $\boldsymbol{\Sigma_{\mathbf{X}}}$, and $\mathbf{U}^\dagger$ denotes the conjugate transpose of $\mathbf{U}$.

The $\alpha$-power of the matrix $\boldsymbol{\Sigma_{\mathbf{X}}}$ is:

\begin{equation}
(\boldsymbol{\Sigma_{\mathbf{X}}})^\alpha = (\mathbf{U} \mathbf{\Lambda} \mathbf{U}^\dagger)^\alpha = \mathbf{U} \mathbf{\Lambda}^\alpha \mathbf{U}^\dagger,
\end{equation}

where $\mathbf{\Lambda}^\alpha$ is also a diagonal matrix with elements $\sigma_i^\alpha$:

\begin{equation}
\mathbf{\Lambda}^\alpha = \text{diag}(\sigma_1^\alpha, \sigma_2^\alpha, \ldots).
\end{equation}

We now calculate the trace $\mathrm{tr}((\boldsymbol{\Sigma_{\mathbf{X}}})^\alpha)$ using the cyclic property of the trace, $\mathrm{tr}(\mathbf{A}\mathbf{B}\mathbf{C}) = \mathrm{tr}(\mathbf{C}\mathbf{A}\mathbf{B})$, and the fact that $\mathbf{U}^\dagger \mathbf{U} = \mathbf{I}$ (since $\mathbf{U}$ is unitary):

\begin{equation}
\begin{split}
    \mathrm{tr}((\boldsymbol{\Sigma_{\mathbf{X}}})^\alpha) &= \mathrm{tr}(\mathbf{U} \mathbf{\Lambda}^\alpha \mathbf{U}^\dagger) \\
    &= \mathrm{tr}(\mathbf{\Lambda}^\alpha \mathbf{U}^\dagger \mathbf{U}) \\
    &= \mathrm{tr}(\mathbf{\Lambda}^\alpha \mathbf{I}) \\
    &= \mathrm{tr}(\mathbf{\Lambda}^\alpha).
\end{split}
\end{equation}

The trace of the diagonal matrix $\mathbf{\Lambda}^\alpha$ is the sum of its diagonal elements:

\begin{equation}
\mathrm{tr}(\mathbf{\Lambda}^\alpha) = \sum_i \sigma_i^\alpha.
\end{equation}

Therefore, we get:

\begin{equation}
S_\alpha(\boldsymbol{\Sigma_{\mathbf{X}}}) = \frac{1}{1-\alpha} \log \left(\sum_i \sigma_i^\alpha\right).
\end{equation}

\subsection{Proof of Lemma 2}
\label{app:proof_lemma2}

To calculate $\lim_{\alpha \to 1} \frac{1}{1 - \alpha} \log \left( \sum_i \sigma_i^\alpha \right)$, considering the Taylor expansion of $\sum_i \sigma_i^\alpha$:

\begin{align}
\sum_i \sigma_i^\alpha &= \sum_i \sigma_i \cdot e^{(\alpha-1)\log \sigma_i} \nonumber \\
&\approx \sum_i \sigma_i \left(1 + (\alpha - 1) \log \sigma_i \right)  \\
&= 1 + (\alpha - 1)\sum_i \sigma_i \log \sigma_i \nonumber.
\end{align}

As $\alpha \rightarrow 1$, we can use the approximation $\log(1+x) \approx x$ for small $x$. Therefore, we get:

\begin{equation}
S_{\alpha}(\boldsymbol{\Sigma_{\mathbf{X}}}) \approx - \sum_i \sigma_i \log \sigma_i.
\end{equation}

\subsection{Proof of Lemma 3}
\label{app:proof_lemma3}

In quantum mechanics, the density matrix $\rho$ serves as a statistical ensemble description of a quantum state, characterizing the inherent uncertainty of the system. 
We demonstrate that the trace-normalized  covariance matrix $\boldsymbol{\Sigma}_{\mathbf{X}}$ functions as the counterpart to $\rho$ within the visual representation space. 
Mathematically, $\boldsymbol{\Sigma}_{\mathbf{X}}$ must be positive semi-definite with a unit trace ($\text{tr}(\rho) = 1$) to serve as a valid density matrix. As a normalized covariance matrix, $\boldsymbol{\Sigma}_{\mathbf{X}}$ inherently satisfies these constraints: it is symmetric and positive semi-definite by definition, while trace normalization guarantees a unit trace.
By definition, $\boldsymbol{\Sigma}_{\mathbf{X}}$ functions as the statistical description of the feature state, essentially playing the role of the density matrix $\rho$ in quantum state space.

The standard definition of Von Neumann entropy is given by $S = -\mathrm{tr}(\rho \log \rho)$. We substitute $\rho$ with $\boldsymbol{\Sigma_{\mathbf{X}}}$ and utilize the spectral decomposition. Since $\boldsymbol{\Sigma_{\mathbf{X}}}$ is a real symmetric matrix, it can be diagonalized as $\boldsymbol{\Sigma_{\mathbf{X}}} = \mathbf{U} \mathbf{\Lambda} \mathbf{U}^T$, where $\mathbf{U}$ is an orthogonal matrix satisfying $\mathbf{U}^T \mathbf{U} = \mathbf{I}$, and $\mathbf{\Lambda} = \text{diag}(\sigma_1, \sigma_2, \dots)$ is the diagonal matrix of the eigenvalues $\sigma_i$.

The matrix logarithm of $\boldsymbol{\Sigma_{\mathbf{X}}}$ is defined via its spectral decomposition:
\begin{equation}
\begin{split}
    \log(\boldsymbol{\Sigma_{\mathbf{X}}}) &= \mathbf{U} \log(\mathbf{\Lambda}) \mathbf{U}^T \\
    &= \mathbf{U} \text{diag}(\log \sigma_1, \log \sigma_2, \dots) \mathbf{U}^T.
\end{split}
\end{equation}

Substituting this into the trace definition:

{ \small
\begin{align}
    -\mathrm{tr}(\boldsymbol{\Sigma_{\mathbf{X}}} \log \boldsymbol{\Sigma_{\mathbf{X}}}) 
    &= -\mathrm{tr}\left( (\mathbf{U} \mathbf{\Lambda} \mathbf{U}^T) \cdot (\mathbf{U} \log(\mathbf{\Lambda}) \mathbf{U}^T) \right) \notag \\
    &= -\mathrm{tr}\left( \mathbf{U} \mathbf{\Lambda} (\mathbf{U}^T \mathbf{U}) \log(\mathbf{\Lambda}) \mathbf{U}^T \right).
\end{align}
}

Using the property $\mathbf{U}^T \mathbf{U} = \mathbf{I}$, this simplifies to:

\begin{equation}
    -\mathrm{tr}\left( \mathbf{U} (\mathbf{\Lambda} \log \mathbf{\Lambda}) \mathbf{U}^T \right).
\end{equation}

By the cyclic property of the trace : $\mathrm{tr}(\mathbf{A}\mathbf{B}\mathbf{C}) = \mathrm{tr}(\mathbf{B}\mathbf{C}\mathbf{A})$, we have:

\begin{align}
    -\mathrm{tr}\left( \mathbf{U} (\mathbf{\Lambda} \log \mathbf{\Lambda}) \mathbf{U}^T \right) \notag
    &= -\mathrm{tr}\left( (\mathbf{\Lambda} \log \mathbf{\Lambda}) \mathbf{U}^T \mathbf{U} \right) \\
    &= -\mathrm{tr}(\mathbf{\Lambda} \log \mathbf{\Lambda}).
\end{align}

Since $\mathbf{\Lambda}$ is a diagonal matrix, $\mathbf{\Lambda} \log \mathbf{\Lambda}$ is also diagonal with elements $\sigma_i \log \sigma_i$. The trace is simply the sum of these diagonal elements:

\begin{equation}
    -\mathrm{tr}(\mathbf{\Lambda} \log \mathbf{\Lambda}) = -\sum_i \sigma_i \log \sigma_i.
\end{equation}

\subsection{Derivation of FLOPs for LLaVA-1.5-7B}
\label{app:derive_flops_llava}

The LLM backbone utilized in LLaVA-1.5-7B is Vicuna-1.5-7B~\citep{zheng2023judging}, which is obtained by fine-tuning LLaMA-2-7B~\citep{touvron2023llama2openfoundation}. Its architectural design encompasses two principal components: the multi-head attention block (MHA module) and the feed-forward network (FFN module). Both modules are followed by an RMS normalization~\citep{zhang-sennrich-neurips19} and a residual connection.

The MHA module transforms the input $\bm{X}$ into query, key, and value matrices ($\bm{Q}, \bm{K}, \bm{V}$) through linear transformations, computes the attention scores, and aggregates the results from multiple heads:
\begin{align}
\bm{Q}=\bm{X}\bm{W}_Q, \bm{K}&=\bm{X}\bm{W}_K, \bm{V}=\bm{X}\bm{W}_V, \label{eq:qkv}
\end{align}
\vspace{-2.5em} 
\begin{align}
\bm{O} &= \operatorname{Attention}(\bm{Q},\bm{K},\bm{V}) \label{eq:attn} \\
       &= \operatorname{softmax}\left(\frac{\bm{Q}\bm{K}^{\intercal}}{\sqrt{d}}\right)\bm{V}, \notag
\end{align}
\vspace{-1.5em} 
\begin{align}
\bm{X} &= \bm{O}\bm{W}_O, \label{eq:o}
\end{align}
where $\bm{W}_Q, \bm{W}_K, \bm{W}_V, \bm{W}_O \in \mathbb{R}^{d \times d}$ denote the learnable parameters, and $d$ represents the hidden dimension.

The FFN module employs the SwiGLU activation function~\citep{shazeer2020gluvariantsimprovetransformer} to expand the intermediate state dimension via gated linear units, followed by a linear projection to yield the output:
\begin{align}
\bm{X} = [ \operatorname{Swish}(\bm{X}\bm{W}_G) \odot (\bm{X}\bm{W}_U) ] \bm{W}_D, \label{eq:ffn}
\end{align}
where $\odot$ denotes the Hadamard product, while $\bm{W}_G, \bm{W}_U \in \mathbb{R}^{d \times m}$ and $\bm{W}_D \in \mathbb{R}^{m \times d}$ are the parameters, with $m$ representing the intermediate dimension of the FFN.

\noindent \textbf{FLOP Analysis.}
We analyze the FLOPs per transformer layer during the prefill stage, adopting the counting methodology detailed in~\citep{llamaflops}.
For the MHA module, the three linear projections (~\cref{eq:qkv}) involve matrix multiplications requiring $6nd^2$ FLOPs, where $n$ is the sequence length. Applying Rotary Positional Embeddings (RoPE) involves element-wise operations accounting for approximately $6nd$ FLOPs.
Regarding the attention mechanism (~\cref{eq:attn}), computing the correlation between $\bm{Q}$ and $\bm{K}$ requires $2n^2d$ FLOPs. The subsequent softmax operation and scaling require roughly $4n^2h$ FLOPs (where $h$ is the number of heads), while the weighted aggregation with matrix $\bm{V}$ incurs another $2n^2d$ FLOPs. Thus, the core attention mechanism consumes a total of approximately $4n^2d+4n^2h$ FLOPs.
The final output projection (~\cref{eq:o}) in the MHA module adds $2nd^2$ FLOPs.

For the FFN module (~\cref{eq:ffn}), the up-projection and gate-projection require $4nmd$ FLOPs. The SwiGLU activation involves element-wise multiplications and the Swish function, contributing $2nm$ FLOPs. The down-projection requires $2nmd$ FLOPs.
Additionally, operations for RMS Normalization (RMSNorm) and residual connections account for $5nmd$ FLOPs.

Aggregating these components, the total FLOPs per layer $F(n)$ can be derived as follows:
\begin{equation}
F(n) = \underbrace{4n^2d}_{\text{Attn Score}} + \underbrace{8nd^2}_{\text{Projections}} + \underbrace{6nmd}_{\text{FFN}} + \mathcal{O}(n).
\end{equation}
Given that the intermediate dimension in LLaMA-2-7B architectures is typically $m \approx \frac{8}{3}d$, the term $6nmd$ approximates to $16nd^2$. Therefore, the total FLOPs can be simplified to:
\begin{equation}
F(n) \approx 4n^2d + 24nd^2.
\end{equation}

\renewcommand{\multirowsetup}{\centering}
\begin{table*}[t]
    \centering
    \setlength{\tabcolsep}{4.5pt}
    \renewcommand{\arraystretch}{1.33}
    \footnotesize
	\centering
    \caption{\textbf{Performance of LLaVA-1.5-7B with EntropyPrune-q and EntropyPrune-k.} The vanilla number of vision tokens is 576. \textbf{Acc.} represents the average accuracy across all benchmarks. \textbf{Rel.} denotes the relative performance retained compared to the original model. The best performance is highlighted in \textbf{bold}, while the second best is \underline{underlined}.}
	\label{tab:llava-abl-qk}
    \begin{tabular}{p{3.0cm}|c c c c c c c c}
        \shline
        \textbf{Method} & \textbf{MMB}  & \textbf{MME} & \textbf{VQA}$^{\text{T}}$ & \textbf{MMB}$^{\text{C}}$ & \textbf{MMstar} & \textbf{Acc.(\%)} & \textbf{Rel.(\%)} & \textbf{FLOPs} \\
        \shline
        \rowcolor{mygray}
        \multicolumn{9}{c}{\textit{Upper Bound, 576 Tokens} \ $\textbf{(100\%)}$}\\
        LLaVA-1.5-7B & 64.6  & 1862 & 58.3 & 58.1 & 33.1 & 56.1 & 100.0 & 8.80T \\ 
        \hline

        \rowcolor{mygray}
        \multicolumn{9}{c}{\textit{Retain 192 Tokens} \ $\fg{(\downarrow 66.7\%)}$} \\
        FastV \texttt{\scriptsize{(ECCV24)}} & 63.2  & 1796 & 56.4 & \underline{57.4} & 31.1 & 54.4 & 97.0 & 3.99T \\ 
        PDrop \texttt{\scriptsize{(CVPR25)}} & 63.3  & 1765 & 56.2 & 56.5 & 30.7 & 53.9 & 96.1 & 3.79T \\
        DART \texttt{\scriptsize{(EMNLP25)}}& 63.7 & 1833 & 57.0  & 57.1  & 31.4 & 54.9 & 97.9 & 3.69T \\
        CDPruner \texttt{\scriptsize{(NIPS25)}}& 63.2 & 1786 & 57.3  & 55.6 & 31.1 & 54.2 & 96.6 & 3.76T \\
        DivPrune \texttt{\scriptsize{(CVPR25)}} & 62.3  & 1769 & 56.6 & 56.2 & 31.5 & 54.0 & 96.1 & 3.73T \\
        Prumerge \texttt{\scriptsize{(ICCV25)}} & 56.3  & 1515 & 50.1 & 51.3 & 22.0 & 46.8 & 83.3 & 3.71T \\
        EntropyPrune-q & \textbf{64.4} & \textbf{1844} & \underline{57.6} & \textbf{57.9} & \underline{31.5} & \textbf{55.5} & \textbf{98.8} & 3.72T \\
        EntropyPrune-k & \underline{64.3} & \underline{1837} & \textbf{57.8} & 57.2 & \textbf{31.7} & \underline{55.3} & \underline{98.6} & 3.72T \\
        \shline
	\end{tabular}
\end{table*}

\section{Detailed Experiment Settings}
\label{app:detail_exp}

\subsection{Benchmarks}
\label{app:benchmark}

\textbf{MMBench} (\textbf{MMB})~\citep{liu2024mmbench}. MMB is a hierarchical benchmark designed to assess the comprehensive capabilities of MLLMs. It employs a circular evaluation strategy and utilizes ChatGPT to ensure robust matching of model predictions, covering varied abilities such as perception and reasoning. \\
\textbf{MMBench-CN} (\textbf{MMB$^{\text{C}}$})~\citep{liu2024mmbench}. MMB$^{\text{C}}$ is the Chinese version of MMB. \\
\textbf{MME}~\citep{fu2024mmecomprehensiveevaluationbenchmark}. MME provides a comprehensive evaluation suite consisting of 14 subtasks divided into perception and cognition categories. It is designed to rigorously test models while avoiding common pitfalls like data leakage and prompt engineering sensitivity. \\
\textbf{TextVQA} (\textbf{VQA$^{\text{T}}$})~\citep{singh2019towards}. VQA$^{\text{T}}$ focuses on the integration of text within images, evaluating the model's ability to comprehend and reason about both the visual and textual information present. The benchmark includes a series of visual question-answering tasks where the model must interpret visual content and read embedded text to respond correctly. \\
\textbf{MMVet}~\citep{yu2024mm}. MMVet evaluates integrated multimodal capabilities, such as recognition, OCR, and spatial awareness. It proposes an LLM-based evaluator for open-ended outputs. \\ 
\textbf{MMstar}~\citep{mmstar2024}. MMstar is an elite vision-indispensable benchmark that focuses on ``hard" samples. It explicitly filters out questions that can be answered by text alone to ensure that the evaluation genuinely reflects the model's visual understanding capabilities. \\
\textbf{AI2D}~\citep{ai2d2016}. AI2D consists of diagrams and corresponding questions, challenging models to parse and reason about diagrammatic structures, arrows, and labels to solve geometry and science problems. \\
\textbf{MMMU}~\citep{yue2023mmmu}. MMMU is a massive multi-discipline multimodal understanding benchmark. It requires expert-level knowledge and reasoning across diverse fields such as art, science, and engineering, testing the breadth and depth of MLLM intelligence.
\\
\textbf{OCRBench} (\textbf{OCR$^{\text{B}}$})~\citep{ocrb2024}. OCR$^{\text{B}}$ is a comprehensive benchmark dedicated to evaluating Optical Character Recognition (OCR) capabilities. It covers various text-related tasks, including text recognition, scene text VQA, and document understanding. \\
\textbf{MSVD-QA and MSRVTT-QA} \citep{xu2017video}. MSVD-QA and MSRVTT-QA are standard benchmarks for Video Question Answering. They evaluate a model's ability to understand temporal dynamics and reason about events, objects, and actions occurring within video clips.

\subsection{Baselines}
\label{app:baseline}

\textbf{FastV}~\citep{chen2024image} leverages the attention value of the last text token to rank the information of visual tokens after the third layer.
\\
\textbf{PDrop}~\citep{xing2024pyramiddrop} introduces a pyramid token dropping strategy. It progressively reduces the number of visual tokens as the network depth increases, based on the observation that deep layers in MLLMs often exhibit high redundancy.
\\
\textbf{DART}~\citep{wen2025stop} selects a subset of pivot tokens and retains the remaining tokens based on low duplication to the pivots to ensure minimal information loss.
\\
\textbf{DivPrune}~\citep{alvar2025divprune} utilizes a diversity-based pruning metric. Instead of relying on attention scores, it selects a subset of tokens that maximize the diversity of information, ensuring that the retained tokens cover the most distinct visual features.
\\
\textbf{CDPruner}~\citep{zhang2026beyond} reformulates token pruning via determinantal point process (DPP) to maximize the conditional diversity of retained tokens based on instruction relevance.
\\
\textbf{PruMerge}~\citep{shang2025prumerge} filters uninformative tokens based on \texttt{CLS} attention value and clusters the rest via key similarity.

\renewcommand{\multirowsetup}{\centering}
\begin{table}[t]
    \centering
    \setlength{\tabcolsep}{7.0pt}
    \renewcommand{\arraystretch}{1.33}
    \footnotesize
    \caption{\textbf{Comparisons with recent visual token pruning methods on LLaVA-1.5-7B.}}
    \label{tab:recent-baselines}
    \begin{tabular}{l|c c c}
        \shline
        \textbf{Method} & \textbf{MME} & \textbf{VQA}$^{\text{T}}$ & \textbf{MMStar} \\
        \shline

        \rowcolor{mygray}
        \multicolumn{4}{c}{\textit{Retain 20\% Tokens} \ $\fg{(\downarrow 80.0\%)}$} \\
        FitPrune
        & 1775 & 55.7 & 30.1 \\
        EntropyPrune
        & \textbf{1782} & \textbf{56.5} & \textbf{31.8} \\

        \hline
        \textbf{Method} & \textbf{MME} & \textbf{MMB} & \textbf{MMStar} \\
        \shline

        \rowcolor{mygray}
        \multicolumn{4}{c}{\textit{Retain 128 Tokens} \ $\fg{(\downarrow 77.8\%)}$} \\
        SparseVLM
        & 1703 & 60.5 & 30.7 \\
        VisionZip
        & 1756 & 60.3 & 29.3 \\
        EntropyPrune
        & \textbf{1780} & \textbf{60.7} & \textbf{31.5} \\
        \shline
    \end{tabular}
\end{table}

\subsection{Implementation Details.}
All experiments are conducted on Nvidia
A6000 GPU. The implementation is carried
out in Python 3.10, utilizing PyTorch 2.1.2, and
CUDA 11.8. All baseline settings follow the
original paper.

\section{Choice of Feature States for Entropy Calculation}
\label{app:abl_sty_qk}

EntropyPrune evaluates the information of each visual token based on the matrix entropy defined in \cref{sec:token_scoring}. In this section, we compare the performance of two variants: EntropyPrune-q, which utilizes query states for token entropy calculation, and EntropyPrune-k, which utilizes key states. 
As shown in \cref{tab:llava-abl-qk}, both variants achieve superior performance across various benchmarks, demonstrating the robustness of matrix entropy as a reliable metric for token information. Specifically, when reducing the vision tokens to 192, our two variants rank first and second in average accuracy, with EntropyPrune-q retaining 98.8\% performance relative to the base model (near-lossless).
For consistency, the final performance reported in \cref{tab:llava15} defaults to using query states.

\section{Comparisons with Recent Visual Token Pruning Methods}
\label{app:recent_baselines}

To further evaluate EntropyPrune against recent visual token pruning methods, we additionally compare it with FitPrune~\citep{ye2025fit}, SparseVLM~\citep{zhang2024sparsevlm}, and VisionZip~\citep{visionzip} on LLaVA-1.5-7B under matched pruning ratios.

The released implementation of FitPrune supports only a fixed pruning ratio, therefore we compare FitPrune and EntropyPrune under the same 80\% pruning ratio. As shown in \cref{tab:recent-baselines}, EntropyPrune consistently outperforms FitPrune across all evaluated benchmarks. We further compare EntropyPrune with SparseVLM and VisionZip when retaining 128 of 576 visual tokens, corresponding to a 77.8\% pruning ratio. EntropyPrune achieves the best performance across all benchmarks. These results demonstrate that EntropyPrune remains competitive against recent pruning methods under different pruning settings.

The MMBench results of EntropyPrune reported in \cref{tab:recent-baselines} differ slightly from those reported in the main paper. The results in the main paper were obtained using the official MMBench evaluation server together with the LLaVA evaluation code. Since the official evaluation server is no longer available, the additional comparisons reported here are evaluated using VLMEvalKit~\citep{duan2024vlmevalkit} for reproducibility.

\begin{figure}[t]
\begin{center}
\centerline{\includegraphics[width=1.0\linewidth]{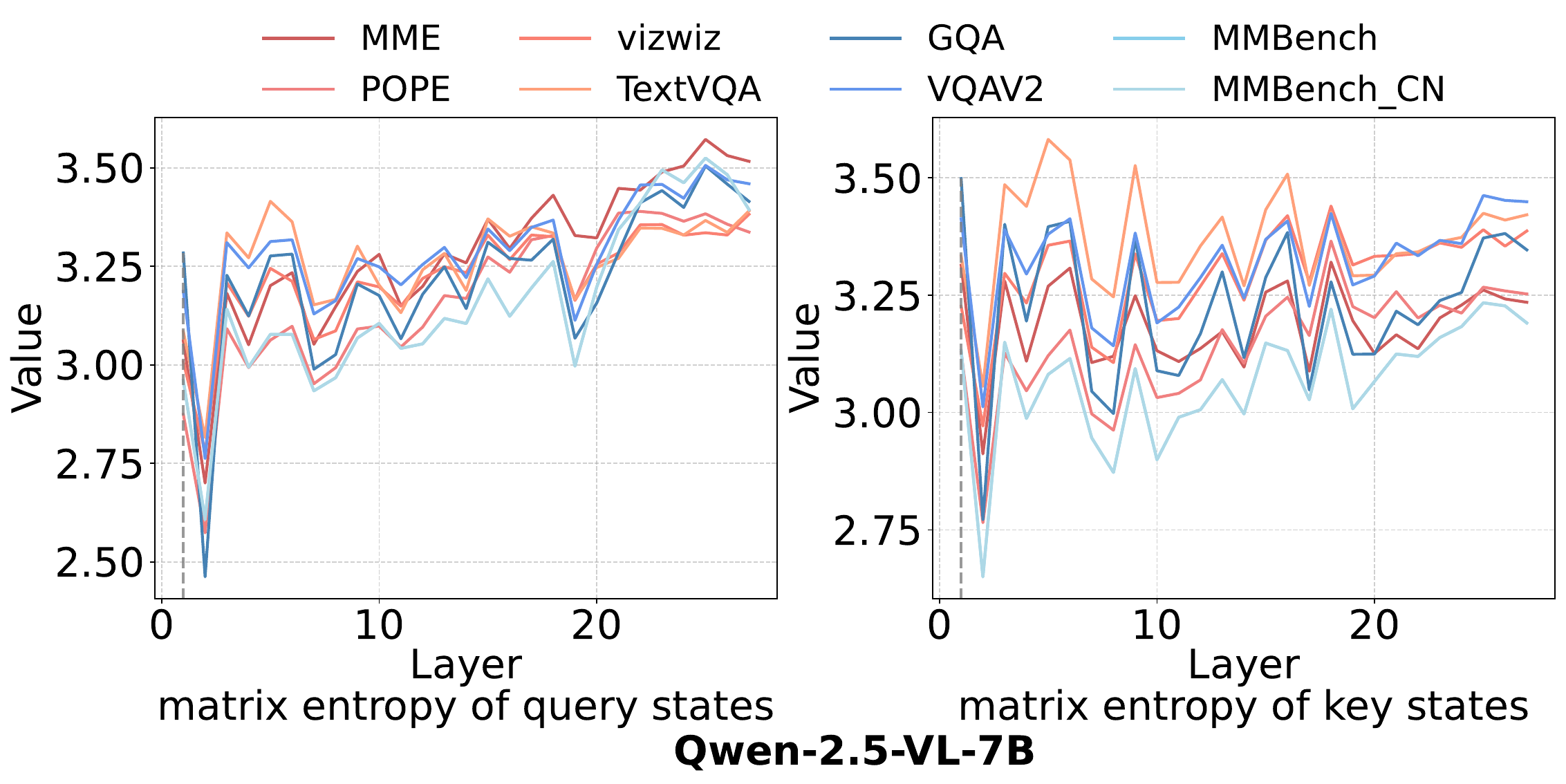}}
\vspace{-4mm}
\caption{
Layer-wise matrix entropy of visual tokens (query and key states) in Qwen-2.5-VL-7B across eight datasets.
A consistent entropy collapse is observed after the first layer across different datasets and feature states, demonstrating that the entropy collapse phenomenon also exists in non-LLaVA architectures.
}
\vspace{-30pt}
\label{fig:layer_matrix_entropy_qwen}
\end{center}
\end{figure}

\section{Generality of Entropy Collapse across Architectures}
\label{app:ecl_generality}

To further examine whether the entropy collapse phenomenon generalizes beyond the LLaVA architecture family, we analyze the layer-wise matrix entropy of visual tokens in Qwen-2.5-VL-7B using randomly sampled inputs from eight datasets, including MME, POPE, VizWiz, TextVQA, GQA, VQAv2, MMBench, and MMBench-CN.

As illustrated in \cref{fig:layer_matrix_entropy_qwen}, matrix entropy exhibit a highly consistent layer-wise pattern across different datasets, independent of query or key states. 
Specifically, the matrix entropy sharply decreases after the first layer. According to our definition, this identifies Layer 1 as the Entropy Collapse Layer (ECL) for Qwen-2.5-VL-7B.

Notably, this behavior closely resembles the entropy dynamics observed in LLaVA-1.5-7B and LLaVA-NeXT-7B, demonstrating that the entropy collapse phenomenon is not specific to the LLaVA architecture family. Moreover, the location of the ECL can vary across different MLLM architectures. These results demonstrate that the entropy collapse is not specific to the LLaVA architecture, but reflects a general phenomenon in the layer-wise evolution of visual token representations in multimodal large language models.

\end{document}